\documentclass{article}

\usepackage{microtype}
\usepackage{graphicx}
\usepackage{subfigure}
\usepackage{amssymb}
\usepackage{siunitx}
\usepackage{booktabs} 
\usepackage{bbm}
\usepackage{color,soul}
\usepackage{multirow}
\usepackage{array}
\usepackage{booktabs}
\usepackage{tabularx}
\newcolumntype{P}[1]{>{\centering\arraybackslash}p{#1}}
\newcolumntype{L}[1]{>{\arraybackslash}p{#1}}
\usepackage{hyperref}

\usepackage[accepted]{mlsys2023}

\mlsystitlerunning{Efficient Vertical Federated Learning with Secure Aggregation}

\begin{document}

\twocolumn[
\mlsystitle{Efficient Vertical Federated Learning with Secure Aggregation}



\mlsyssetsymbol{equal}{*}

\begin{mlsysauthorlist}
\mlsysauthor{Xinchi Qiu}{equal,cam}
\mlsysauthor{Heng Pan}{equal,cam}
\mlsysauthor{Wanru Zhao}{cam}
\mlsysauthor{Chenyang Ma}{cam}
\mlsysauthor{Pedro Porto Buarque de Gusmão}{cam}
\mlsysauthor{Nicholas D. Lane}{cam}
\end{mlsysauthorlist}

\mlsysaffiliation{cam}{Camputer Laboratory, University of Cambridge, UK}

\mlsyscorrespondingauthor{Xinchi Qiu}{xq227@cam.ac.uk}
\mlsyscorrespondingauthor{Heng Pan}{ac.panh99@gmail.com}

\mlsyskeywords{Machine Learning, Security, Privacy, Security Aggregation}

\vskip 0.3in

\begin{abstract}
The majority of work in privacy-preserving federated learning (FL) has been focusing on horizontally partitioned datasets where clients share the same sets of features and can train complete models independently. However, in many interesting problems, such as financial fraud detection and disease detection, individual data points are scattered across different clients/organizations in vertical federated learning. Solutions for this type of FL require the exchange of gradients between participants and rarely consider privacy and security concerns, posing a potential risk of privacy leakage. 
In this work, we present a novel design for training vertical FL securely and efficiently using state-of-the-art security modules for secure aggregation. We demonstrate empirically that our method does not impact training performance whilst obtaining \num{9.1e2} $\sim$ \num{3.8e4} speedup compared to homomorphic encryption (HE).

\end{abstract}
]



\printAffiliationsAndNotice{\mlsysEqualContribution} 

\section{Introduction}\label{sec:intro}


Federated Learning (FL) is a machine learning paradigm that enables the training of a global model using decentralized datasets without requiring sharing of raw and sensitive data from participating parties \cite{fedavg, fedprox,wei2022vertical}. Under FL, individual institutions or devices train a common global model collaboratively, agreeing on a possible third party or central server to orchestrate the learning and perform model aggregation. 

In terms of data partitioning, FL can be categorized as either horizontal or vertical scenarios. Most existing works focus on horizontal FL (HFL), which requires all participants to use the same feature space but different sample spaces \citet{yang2019federated, wei2022vertical, vfl}. This partitioning scheme is usually found in the cross-device setup where clients are often mobile or IoT devices with heterogeneous datasets and resources under complex distributed networks \cite{yu2021toward, qiu2021zerofl}. Under Vertical FL (VFL), however, data points a partitioned across clients. This is often found in cross-silo setups where participating clients, such as hospitals and research institutions, may hold complementary pieces of information for the same data points. For example, two different hospitals may hold different medical data for the same patient.


The need for VFL has arisen massively in the industry these years \cite{vfl, liu2020asymmetrical}. For example, a financial institution would like to train a financial crime detection model, but it only has limited features in its own institution that limit the performance. The institution would like to have access to more private information, such as account information, that various banks might have. However, this is a deal breaker for financial applications as such data is very sensitive, and legal restrictions (e.g., GDPR) can prevent it from being shared across institutions. With VFL, institutions and companies that own only small and fragmented data have constantly been looking for other institutions to collaboratively develop a shared model for maximizing data utilization \cite{li2021survey}. 

Due to their different data structures, training procedures for HFL and VFL can be very different. Each client in HFL trains a complete copy of the global model on their local dataset and sends model updates to a centralized server for aggregation. 
Under VFL,  each client, holding certain features of the whole dataset, contributes to a sub-module of the global model. This means that intermediate activations or gradients need to be shared between clients during the training process, posing a potential risk for privacy leakage, as the original raw data can be reconstructed from said gradients~\cite{dlg, idlg, jin2021cafe}. While most research has focused on designing methods to train the global model under VFL better, fewer efforts have been devoted to providing a secure way of training. The method proposed by~\cite{liu2020asymmetrical} only tries to protect the sample IDs, rather than all the raw private data; \cite{chen2020vafl} perturbed local embedding to ensure data privacy and improve communication efficiency, which has strict requirements for the embedding and can impact the overall performance. There are also BlindFL \cite{fu2022blindfl}, ACML \cite{zhang2020acml}, and PrADA \cite{ kang2022prada} are homomorphic encryption (HE) based solutions. These approaches often incur significant communication and computation overheads. Moreover, their fixed design cannot be extended to multiple-party scenarios.


Our work aims to provide an efficient and privacy-preserving way of training under the Vertical FL setup. We begin by describing in Section \ref{sec:problem_setup} the specific problem setup. We then describe a secure aggregation method tailored to solve this kind of FL problem in Section~\ref{sec:secagg}. Finally, we demonstrate its applicability with extensive experiments described in Section \ref{sec:exp} to show that our SA incurs minimal overhead. In addition, we demonstrate that our method does not impact training performance whilst obtaining \num{9.1e2} $\sim$ \num{3.8e4} speedup compared to HE.

\section{Problem Setup}\label{sec:problem_setup}
In this section, we formally define the problem of VFL for classification. 
We consider a $C$ class classification problem defined over a compact space $\mathcal{X}$ and a label space $\mathcal{Y} = [C]$, where $[L] = \{ 1,...,C \}$. 

Following the setup as in previous literature \cite{vfl}, we define two kinds of clients in the vertical FL settings. The first type is the \emph{active party}, which holds all the samples with the ground-truth label and multiple features, and there usually is only one active party. The second type is called the \emph{passive parties}, which only holds some features that are not overlapping with the features in the active party. 
Let Client 0 ($\mathcal{C}_0$) denote the active party with features $x_{1,...,m}$, and the rest be the passive parties ($\mathcal{C}_i$ with $i= 1,...,N$) with features $x_{m,...,n}$.
Passive parties can be clustered by the feature set they owned. Multiple passive parties can hold different samples with the same feature set.
Let $f$ be the function for the neural network parameterized over the hypothesis class $w$, which is the weight of the neural network. $\mathcal{L}(\textbf{w})$ is the loss function, and we assume the widely used cross-entropy loss.

Since the active party holds the ground-truth label and some features, it is capable of training the model and making inferences only using its own local data. The aim of training using VFL is to incorporate other features that exist in the passive parties to boost performance while maintaining the privacy of the data features in either active or passive parties.   

This setup can be commonly found in real-world scenarios. For example, for the financial crime detection task, each financial institution will have different features regarding account information or transaction information. However, for financial applications, such data is very sensitive, and legal restrictions (e.g., GDPR) can prevent it from being shared across institutions. Another example will be in the commercial ad ranking systems, in the sense that each company or organization might have different information for the same customer, but the label (the click rate) will only be stored in the application platform. 

\begin{figure}[t]
    \small
    \centering
    \includegraphics[width=0.35\textwidth]{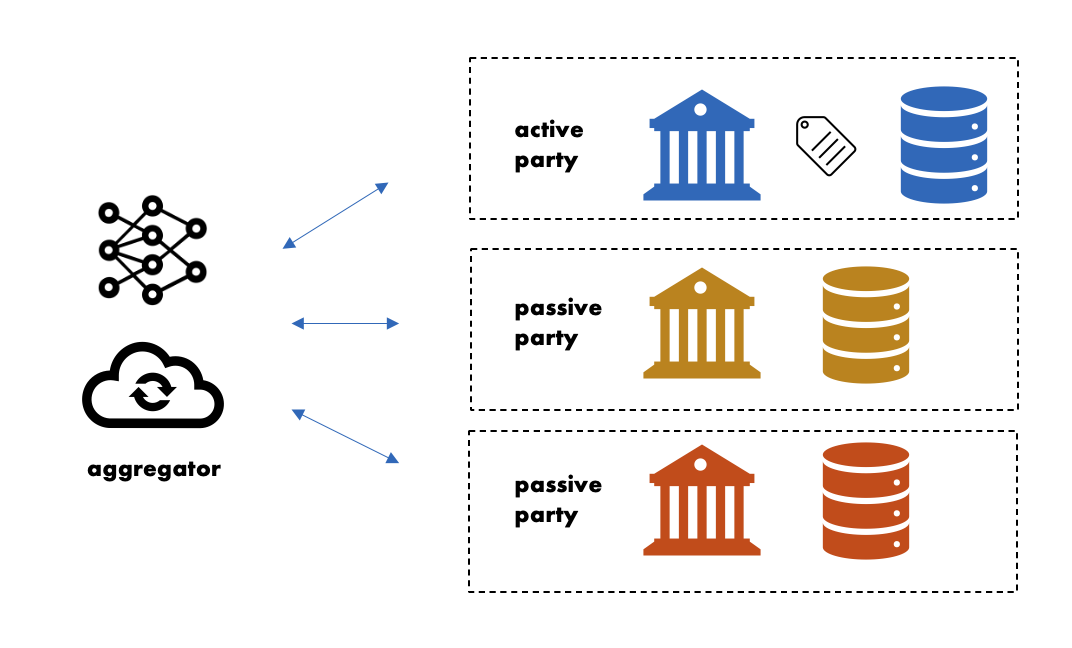}

    \caption{Illustration of the VFL problem setup. Only the active party has the label; each party might have different features.}

    \label{fig:setup}
\end{figure}

\section{centralized solution}\label{sec:centralized}
This section describes the training method for the problem setup explained in Section \ref{sec:problem_setup}.

In a horizontal FL scenario, where datasets are horizontally partitioned, FL solutions are typically derived from a centralized one, which in turn provides an upper-bound target efficiency. However, this approach is not well suited in the case of vertically-partitioned FL, where different feature types are distributed across different participants.
 
In order to provide a realistic upper-bound target for the VFL solution, the centralized solution is designed to impose these data access restrictions from its conception. In this section, we describe in detail how to train and make inferences for the centralized model tailored to VFL. The training can be divided into two steps as explained below.

\textbf{Pre-training:} The first stage is the pre-training phase. It can be pre-training using features and the label in the active party or the feature embedding extraction in both active and passive parties. Let $f_i$, $i= 1,...,k$ be the embedding extracted from the active party and $f_i$, $i= k,...,l$ be all the embedding from passive parties. Noted that the embedding from the passive party can be from different clients.

\textbf{Neural Network with VFL:} After the pre-training step, all embeddings are then served as the input together to the neural network to make the final prediction. During inference time, the prediction can also be obtained following these two steps. 




\section{Secure Aggregation Method}
\label{sec:secagg}
In this section, we detail our secure aggregation method for VFL. We incorporate two different methods to ensure the privacy and security of the private sensitive information in local datasets that existed on the client side. First, we use encryption to help the mini-batch selection \emph{without revealing account information to any third party not holding the account information}. Second, we adapt the idea of \emph{secure aggregation} \cite{secagg} for the gradient aggregation during the optimization steps. The procedures can be divided into three phases: setup phase (Section \ref{sec:setup}), training phase (Section \ref{sec:trainingphase}), and testing phase (Section \ref{sec:testingphase}). 

Since features are distributed across different clients, both the forward pass and the backward pass cannot be computed at the same place at the same time. Therefore, we also explain the setup phase for the key exchange in Section \ref{sec:setup} and the training procedure in detail regarding the mini-batch selection, forward pass, and backward pass below during the training phase in Section \ref{sec:trainingphase}.

\subsubsection{Setup phase} \label{sec:setup}

Similar to the centralized solution, the first step happens at the active client using only the active dataset to train a pre-train model to obtain the first stage prediction. The prediction estimation will later be used with the features from the passive parties to train the neural network. 


\textbf{Key generating step:} The first step of secure aggregation is to generate shared secrets. We use the Elliptic-curve Diffie-Hellman (ECDH) key agreement protocol \cite{diffie-hellman, ecdh} to generate shared secrets through insecure channels between all clients. The shared secrets will be used to build secure pairwise channels by symmetric encryption and facilitate secure aggregation. During the setup phase, the central aggregator requests public keys from all participating clients. Then, $\forall i$, Client $i$ generates one pair of secret key $sk_{i}^{(j)}$ and public key $pk_i^{(j)}$ for each Client $j$ and sends public keys to the aggregator. $\forall i \ne j, pk_i^{\left(j\right)}$ is then forwarded to Client $j$.
Once received, Client $i$ and Client $j$ can generate a shared secret $ss_{ij}=ss_{ji}$ from $(sk_i^{(j)}, pk_j^{(i)})$ or $(sk_j^{(i)}, pk_i^{(j)})$.

\subsubsection{Training Phase}\label{sec:trainingphase}

This section explains the training phase in detail.

\paragraph{Mini-batch selection} We assume that the active party knows which passive parties hold the features of a given sample. This can be realized by Private Set Intersection \cite{lu2020psi, zhou2021psi}. We denote the identifier for samples as sample ID, which is shared among all parties. Since the active party has the information regarding each sample and the ground truth label, the mini-batch selection will start from the active party. It will first select a batch of data in the active party ($\mathcal{C}_0$). The sample ID will be encrypted using $ss_{0\,i}$ as key if the partial features of the sample are held by passive party $\mathcal{C}_i$. 
The active party will then upload the encrypted ID batch to the aggregator, which will broadcast it to all passive parties.  As sample IDs are encrypted using different keys, \emph{each passive party can only decrypt sample IDs existing in its dataset}, which prevents any party from knowing extra information about the batch and samples in it.

\paragraph{Forward pass} 
After the pre-training steps in the active party, each sample will have embeddings $f_{i}, i= 1,...,k$, which is then fed into the neural network together with the features embeddings  $f_{k,...,l}$ from passive parties. 

\begin{equation} \label{eqn:pred}
    \hat{y} = f(w;f, i=1,...,l)
\end{equation}

In the first round after the setup phase, the active party will initialize the model parameters, i.e., $w_{t=0}$. During the forward pass, the active party will send to the aggregator the encrypted batch, the ground-truth labels of the selected batch, and the initialized model parameters $w_{t=0}$. We assume it is safe to share the labels, since without any additional information, such as sample ID, adversaries cannot reveal any sensitive information by only knowing the labels. 


Likewise, once received model weights and the encrypted batch are, passive party $p$ will try decrypting each value in the batch and then reply with the following masked vector:

\begin{equation}
    \Big( \sum_{i=k}^{l} \mathbbm{1}(f_i^{(j)} \in D_p) w_i f_i^{(j)}
    \Big)_{j=1}^{B}
    + \mathbf{n}_p
\end{equation}
where $D_p$ is the dataset of passive party $p$ and only sums over the features that exist in the local dataset. $\mathbf{n}_p$ is a uniformly random vector. Notes that the active party will also send the masked activation to the aggregator to finish the forward pass.

Following the idea of \emph{Secure Aggregation}, we make added noises cancel out each other, i.e., $\sum_p{\mathbf{n}_p} = \mathbf{0}$, which is the summation of a series of random numbers generated by a Pseudo-Random Generator (PRG) that can generate sequences of uniformly pseudo-random numbers given a seed, as shown in Equation \ref{eqn:noise} and \ref{eqn:sum_noise}.

\begin{equation} \label{eqn:noise}
    \mathbf{n}_i = -\sum_{j < i}{\textsc{PRG}(ss_{ij})} + \sum_{j > i}{\textsc{PRG}(ss_{ij})}
\end{equation}
\begin{equation} \label{eqn:sum_noise}
    \sum_i{\mathbf{n}_i} = \sum_i{\sum_{j>i}{\big(
    \textsc{PRG}(ss_{ij}) - \textsc{PRG}(ss_{ji})\big)}} = \mathbf{0}
\end{equation}

Through the chosen added noises, after receiving all masked vectors from all passive clients, the aggregator can compute output accurately without knowledge of any individual value from each client by adding them together:

\begin{equation}
    \big(z^{(k)}\big)_{k=1}^B = \Big( \underbrace{\sum_{i=1}^{k} w_i f_i^{(k)}}_{active}  + \underbrace{\sum_{i=k}^{l} w_i f_i^{(k)}}_{passive} \Big)_{k=1}^B
    + \underbrace{\sum_p{\mathbf{n}_p}}_{noise}
\end{equation}
where $\sum_p{\mathbf{n}_p} = \mathbf{0}$.

Due to the fact that all activation are masked using random noise, any party cannot reveal additional information about other features existing in other clients, even if the aggregator colludes with one or multiple passive clients. With $\big(z^{(k)}\big)_{k=1}^B$, the aggregator can then compute the output.

\paragraph{Backward pass}

After receiving the label, each client p can compute the partial gradient $\mathcal{L}(f_i)$ for the mini-batch with respect to the features that the local data exists. Then, each client will send back to the aggregator the masked gradient with mask noise $\mathbf{n}_p$ like in the forward pass. The indicator function is to choose the features that exist in the local dataset $D_p$. Therefore, the formula can be found below:





\begin{equation}
    \Big( \sum_{k=1}^B \mathbbm{1} (f_i^{(k)} \in D_p) \nabla \mathcal{L}(f_i^{(k)})\Big)_{i=1}^{l}  + 
    \mathbf{n}_{p}
\end{equation}

Similarly, the aggregator can compute the summation of masked vectors from passive clients and send the result to the active client, which can then compute the aggregate gradient with respect to each model parameter. It is worth noting that the aggregator only obtains a masked vector,  so $\frac{\partial}{\partial w_0}l$ can only be computed locally at the active party. Thus, the summed batch gradients are only visible to the active party, and any individual gradient is kept from any participating party to protect the sensitive data leakage through the individual gradient.




\subsubsection{Testing Phase}\label{sec:testingphase}
During the testing phase, the active party will first send the encrypted batch information and the masked vector $\big(\sum_{i=1}^{k} w_i f_i^{(k)}
\big)_{k=1}^{B} + \mathbf{n}_0$ to the aggregator. After receiving the encrypted batch information, the encrypted batch information is shared with the passive clients. Then each passive client $p$ computes its masked vector $\Big( \sum_{i=m}^{n} \mathbbm{1}(x_i^{(k)} \in D_p) w_i f_i^{(k)}\Big)_{k=1}^{B} + \mathbf{n}_p$ and sends it back to the aggregator. Lastly, after the aggregator receives all values, it can make the prediction. 

\section{discussion}\label{sec:diss}
\subsection{Threat-model and Privacy Guarantee}\label{sec:diss1}
We consider a \emph{threat model} where both clients and the server are \emph{honest-but-curious}, i.e.\ they are expected to follow the pre-defined training protocol whilst trying to learn as much information as possible from the models they receive. To reduce the risk of information leakage, our solution used state-of-the-art security modules for \emph{Secure Aggregation} by adding the masked gradient when communicating with the central aggregator.

Secure Aggregation is achieved through the use of masking and encryption of gradients before they are sent back to the aggregator. It can thus prevent the aggregator from using the received model update to gain knowledge about the sensitive data on the client side. By using a masked gradient, it is impossible to reconstruct the original data or make inferences about the sensitive information in the dataset without knowing the mask. Therefore, our method protects the model from both data reconstruction and membership inference attacks.

In addition, while our current implementation could be vulnerable to active adversaries, our FL solution can be extrapolated very easily to include \emph{malicious} settings by introducing a public-key infrastructure (PKI) that can verify the identity of the sender \cite{bonawitz2017secagg}. It can thus be further protected from malicious attacks. 

Although our method does not allow exposing secret keys to other parties and demonstrates robustness against collusion between the aggregator and passive parties, in practical settings, the risk of secret key leakage persists. Thus, for the sake of privacy, it is necessary to routinely regenerate keys for symmetric encryption and SA, specifically, by executing the setup phase after every K iteration, in both the training and testing stages. The value of K can vary in real-world scenarios, but the larger value will inevitably incur higher risks of keys being compromised. In the event of key leakage, an attacker will only have access to a limited amount of information instead of all encrypted information if keys are regenerated periodically.

\subsection{Other Considerations}
Scalability in FL directly defines the maximum number of participating clients in the system and indirectly dictates how much data will be used during training and how generalizable the trained model will be. This can be severely hampered if the individual privacy modules or the underlying FL framework are limited on the number of clients participating or the particular way of the data partition. Our core solution is agnostic on the number of participating clients and the data partition schemes, especially in the cross-silo scenarios. As a result, our solution's scalability is only dependent on the underlying FL framework and on how key generation and key exchange between clients are handled. Also, our solution is scalable in the sense that, unlike homomorphic encryption, we employ lightweight masks through random noise, and the mask can be naturally decrypted through summation. 




\section{Experiments}\label{sec:exp}
We conducted extensive experiments on three classification datasets. Federated learning is simulated with the Virtual Client Engine of the Flower toolkit \citep{beutel2020flower}.

\subsection{Datasets}
Experiments are conducted over three datasets: \emph{Banking dataset} \cite{bankdataset}, \emph{Adult income dataset} \cite{adultincome} and \emph{Taobao ad-display/click dataset} \cite{taobao}. The banking dataset is related to the direct marketing campaigns of a Portuguese banking institution. It contains $45,211$ rows and $18$ columns ordered by date. The adult income dataset is a classification dataset aiming to predict whether the income exceeds 50K a year based on census data. It contains $48,842$ and $14$ columns. We also conduct our experiment over a production scale ad-display/click dataset of Taobao \cite{taobao}. The dataset contains $26$ million interactions (click/non-click when an Ad was shown) and $847$ thousand items across an 8-day period. 


\subsection{Feature and Client Partitioning} 
\textbf{Banking Datasets:} We keep the \texttt{housing}, \texttt{loan}, \texttt{contact}, \texttt{day}, \texttt{month},
\texttt{campaign}, \texttt{pdays}, \texttt{previous}, \texttt{poutcome} features in the \emph{active party}. Features \texttt{default}, \texttt{balance} are seen in \emph{passive parties} $1$ and $2$, while \texttt{age}, \texttt{job}, \texttt{marital}, \texttt{education} are kept in \emph{passive parties} $3$ and $4$.

\textbf{Adult Income Dataset:} We keep features \texttt{workclass}, \texttt{occupation}, \texttt{capital-gain}, \texttt{capital-loss}, \texttt{hours-per-week} in the active party and \texttt{race}, \texttt{marital-status}, \texttt{relationship}, \texttt{age} \texttt{gender}, \texttt{native-country} are kept by \emph{passive parties} $1$ and $2$, while \texttt{education} is held by \emph{passive parties} $3$ and $4$.

\textbf{Taobao Dataset:} We keep \texttt{pid}, \texttt{cms\_group\_id}, \texttt{final\_gender\_code}, \texttt{age\_level}, \texttt{pvalue\_level}, \texttt{shopping\_level}, \texttt{occupation}, \texttt{cate\_id}, \texttt{brand}, \texttt{new\_user\_class\_level }, \texttt{price} features in the active party and \texttt{final\_gender\_code}, \texttt{age\_level}, \texttt{occupation} are possessed by \emph{passive parties} $1$ and $2$, while \texttt{pvalue\_level}, \texttt{shopping\_level} are kept in the \emph{passive parties} $3$ and $4$.


\textbf{Model Architecture.} 
Features and models are partitioned among different parties in experimental settings. 
In the Banking dataset, the active party used Linear(57, 64); passive party $1$ and $2$ used unbiased Linear(3, 64); passive party $3$ and $4$ used unbiased Linear(20, 64). The three local modules combined are equivalent to Linear(80, 64). The global module owned by the aggregator comprised Linear(64, 1).
In the Adult Income dataset, the active party, passive party $1$ and $2$, and passive party $3$ and $4$ possessed Linear(27, 64), unbiased Linear(63, 64), and unbiased Linear(16, 64) respectively. The three are equivalent to Linear(106, 64). The global module had Linear(64, 1).
In the Taobao dataset, Linear(197, 128), Linear(11, 128), and Linear(6, 128), which were equivalent to Linear(214, 128), were utilised by the active party, passive party $1$ and $2$, and passive party $3$ and $4$ respectively. The aggregator maintained a global module with Linear(128, 1).
We used a learning rate of 0.01 and a batch size of 256. We applied ReLU activation to all layers except the output layer. 



\subsection{Compute and Communication Overhead}
\label{sec:overhead}
We conduct experiments over three datasets to measure both the computation and the communication cost of VFL training. The computation cost is measured through CPU time (in milliseconds), and the communication cost is measured through the transmission size (in bytes). We also measure the overhead cost that shows the extra CPU time or communication compared to unsecured VFL training. All experiments are reported with 1 setup phase and 5 training rounds, and each experiment is repeated 10 times, and averages and standard deviations are reported. Each experiment is repeated 10 times.

As mentioned in Section \ref{sec:diss1}, in practice, each party should create new key pairs routinely to mitigate the risk of adversaries from accessing confidential information in the event of secret key leakage. In our experiments, the key pairs and the shared secrets will be regenerated for every 5 iterations.

\subsection{Results}\label{sec:res}

\begin{table*}[h!]
\centering
\begin{tabular}{c|P{1.6cm}P{1.4cm}P{1.4cm}P{1.4cm}|P{1.4cm}P{1.4cm}P{1.4cm}P{1.4cm}}
\toprule
\multicolumn{1}{c}{} & \multicolumn{4}{c}{Active Party CPU time (ms)} & \multicolumn{4}{c}{Passive Party CPU time (ms)} \\
\cmidrule(lr){2-5} \cmidrule(lr){6-9}
\multicolumn{1}{c}{} & \multicolumn{2}{c}{Training phase} & \multicolumn{2}{c}{Testing phase} & \multicolumn{2}{c}{Training phase} & \multicolumn{2}{c}{Testing phase} \\
\cmidrule(lr){2-3} \cmidrule(lr){4-5} \cmidrule(lr){6-7} \cmidrule(lr){8-9}
\multicolumn{1}{c|}{Dataset} & Total & Overhead & Total & Overhead & Total & Overhead & Total & Overhead \\
\midrule
Banking    & $1162\pm527$ & $198\pm12$ & $325\pm15$ & $197\pm12$ & $152\pm6$ & $116\pm7$ & $139\pm6$ & $114\pm7$ \\
Adult Income  & $814\pm496$ & $202\pm9$ & $292\pm12$ & $200\pm10$ & $165\pm14$ & $120\pm13$ & $148\pm16$ & $118\pm13$ \\
Taobao  & $2007\pm649$ & $185\pm3$ & $429\pm7$ & $184\pm3$ & $142\pm9$ & $106\pm3$ & $127\pm5$ & $105\pm3$ \\
\bottomrule
\end{tabular}
\caption{Results on the CPU time using secure aggregation on VFL. The CPU times (in milliseconds) are reported. The overhead columns show extra CPU time compared with unsecured VFL training. }
\label{tab:cpu_time}
\end{table*}

\begin{table*}[h!]
\centering
\begin{tabular}{c|P{1.6cm}P{1.4cm}P{1.4cm}P{1.4cm}|P{1.4cm}P{1.4cm}P{1.4cm}P{1.4cm}}
\toprule
\multicolumn{1}{c}{} & \multicolumn{4}{c}{Active Party Data Transmission (bytes)} & \multicolumn{4}{c}{Passive Party Data Transmission (bytes)} \\
\cmidrule(lr){2-5} \cmidrule(lr){6-9}
\multicolumn{1}{c}{} & \multicolumn{2}{c}{Training phase} & \multicolumn{2}{c}{Testing phase} & \multicolumn{2}{c}{Training phase} & \multicolumn{2}{c}{Testing phase} \\
\cmidrule(lr){2-3} \cmidrule(lr){4-5} \cmidrule(lr){6-7} \cmidrule(lr){8-9}
\multicolumn{1}{c|}{Dataset} & Total & Overhead & Total & Overhead & Total & Overhead & Total & Overhead \\
\midrule
Banking    & 959702 & 144826 & 597762 & 144826 & 823803 & 135541 & 464243 & 135541 \\
Adult Income  & 1031382 & 144826 & 597762 & 144826 & 895483 & 135541 & 464243 & 135541 \\
Taobao  & 1629142 & 144826 & 925442 & 144826 & 1493243 & 135541 & 791923 & 135541 \\
\bottomrule
\end{tabular}
\caption{Results on the communication both in size (bytes) using secure aggregation on VFL. The overhead columns show the extra CPU time compared with unsecured VFL training. }
\label{tab:communication}
\end{table*}

In Table \ref{tab:cpu_time} we report the CPU time as a measure to show the computation cost using secure aggregation on VFL. The CPU time is reported in milliseconds and it is reported separately for the active party and passive parties. Table \ref{tab:communication} shows the transmission size in bytes for the method and is also demonstrated on both the active and passive parties. 

As demonstrated in both tables, the overhead accounts for a relatively small part of the total amount, in both CPU time and communication size. The CPU time overhead is caused by parties adding masks to their original output and encryption/decryption of sample IDs. The communication overhead is introduced by broadcasting encrypted sample IDs, which are larger than plain text. As the masks can be cancelled out by summing them, the unmasking process is very efficient. The results are from simple models, and the total amount in practice can dwarf the overhead if more complex models are used.

\subsection{Abalation Study}

In this section, we demonstrate the efficiency of our SA through an ablation study to compare our method and homomorphic encryption (HE). The experiment compares how secure aggregation (SA) and HE process dot productions. 

\begin{figure}
    \centering
    \includegraphics[width=0.44\textwidth]{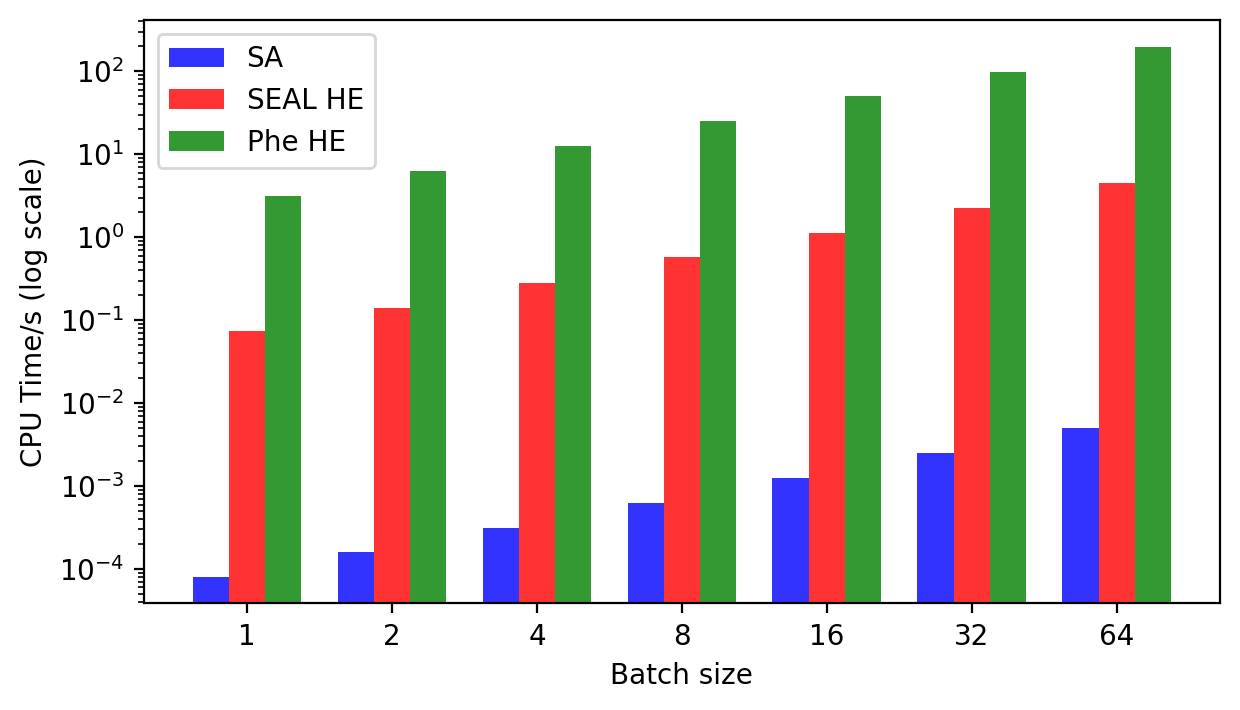}
    \caption{Comparison of average CPU time for different batch sizes, using SA and HE from Phe and SEAL-Python. Y-axis is in log-scale. The results are collected from 10 experiments.}
    \label{fig:cmp}
\end{figure}

Assume the input tensor is of size (Batch size, 8), and the weight tenor is (8, 8). Tensor shapes in the comparison are smaller than tensors used by a passive party in experiments. Given that the HE libraries do not support matrix operations, both SA and HE implementations are not optimized by any Python modules, such as numpy. We use HE functions from Python module \emph{Phe} \cite{phe} and \emph{SEAL-Python} \cite{sealpy}. \emph{Phe} module implements the Pallier cryptosystem in Python, and \emph{SEAL-Python} creates Python bindings for APIs in Microsoft SEAL \cite{sealcrypto} using Pybind11 \cite{pybind11}. The HE implementation in this comparison inevitably involves nested Python loops. For large matrices or more advanced operations, implementations in other languages, e.g., C++, C\#, are more suitable. The results can be found in Fig. \ref{fig:cmp}, which clearly shows the efficiency of SA. It shows that our SA can achieve \num{9.1e2} $\sim$ \num{3.8e4} speedup compared to HE.

\vspace{-4mm}
\section{Conclusion}
\label{sec:conc}
In this work, we consider the challenge of privacy-preserving training in vertical federated learning settings. We provide the first framework to use secure aggregation in the setting of vertical FL by implementing state-of-the-art security modules for Secure Aggregation (SA) by adding noises when communicating with the central aggregator. Our method is efficient and accurate in the sense that it will not change the underlying results and performance by adding security modules. We also provide a unique ablation study between our method and the homomorphic encryption method (HE) to show that our method can achieve \num{9.1e2} $\sim$ \num{3.8e4} speedup compared to homomorphic encryption (HE). Our current method works with the pre-training step. A possible avenue for future work would be to explore how to generalize the secure aggregation method to include all kinds of vertical FL settings.

\nocite{langley00}

\bibliography{bib}
\bibliographystyle{mlsys2023}

%


\end{document}